# Indoor Space Recognition using Deep Convolutional Neural Network: A Case Study at MIT Campus

**Authors**: Fan Zhang*[1,3], Fabio Duarte[1,2], Ruixian Ma[1], Dimitrios Milioris[1], Hui Lin[3,4,5], Carlo Ratti[1]

1. SENSEable City Laboratory,
Massachusetts Institute of Technology,
77 Massachusetts Avenue, Cambridge, MA 02139 USA
{zhangfan, fduarte, milioris, ratti}@mit.edu
2. Pontifícia Universidade Católica do Paraná
3. Shenzhen Research Institute, The Chinese University of Hong Kong,
2nd Yuexing Road, Nanshan District, Shenzhen 518057, China
{humingyuan, huilin}@cuhk.edu.hk
4. Institute of Space and Earth Information Science, The Chinese University of Hong Kong,
Fok Ying Tung Remote Sensing Building, CUHK, ShaTin, N.T., Hong Kong
zhangfan@link.cuhk.edu.hk
5. Department of Geography and Resource Management, The Chinese University of Hong Kong,
Shatin, N.T., Hong Kong

**Correspondence**: [*] Corresponding author. E-mail: zhangfan@link.cuhk.edu.hk

# Abstract

Global Position Systems and other navigation systems that collect spatial data through an array of sensors carried on by people and distributed in space have changed the way we navigate complex environments, such as cities. However, indoor navigation without reliable GPS signals relies on wall-mounted antennas, WiFi, or quantum sensors. Despite the gains of such technologies, underlying these navigation systems is the dismissal of the human wayfinding ability based on visual recognition of spatial features. In this paper, we propose a robust and parsimonious approach using Deep Convolutional Neural Network (DCNN) to recognize and interpret interior space. DCNN has achieved incredible success in object and scene recognition. In this study we design and train a DCNN to classify a pre-zoning indoor space, and from a single phone photo to recognize the learned space features, with no need of additional assistive technology. We collect more than 600,000 images inside MIT campus buildings to train our DCNN model, and achieved 97.9% accuracy in validation dataset and 81.7% accuracy in test dataset based on spatial-scale fixed model. Furthermore, the recognition accuracy and spatial resolution can be potentially improved through multiscale classification model. We identify the discriminative image regions through Class Activating Mapping (CAM) technique, to observe the model's behavior in how to recognize space and interpret it in an abstract way. By evaluating the results with misclassification matrix, we investigate the visual spatial feature of interior space by looking into its visual similarity and visual distinctiveness, giving insights into interior design and human indoor perception and wayfinding research. The contribution of this paper is threefold. First, we propose a robust and parsimonious approach for indoor navigation using DCNN. Second, we demonstrate that DCNN also has a potential capability in space feature learning and recognition, even under severe appearance changes. Third, we introduce a DCNN based approach to look into the visual similarity and visual distinctiveness of interior space.

**Keywords:**

# 1.    Introduction

## 1.1 Background

Global Positioning Systems (GPS) and other navigation systems that collect spatial data through an array of sensors carried on by people and distributed in space, have changed the way we navigate complex environments, such as cities. As Roger McKinlay [1] puts, "the days of being lost should be over".

Wayfind experiences using digital position systems abound. In urban contexts, using real-time mapping tools embedded in mobile devices, it is possible to match the user location with public transportation, traffic and pollution levels, and nearby friends based on social media activity. With this information, applications propose less polluted and faster routes, or those with higher possibilities of social interactions. In these cases, "the map emerges through contingent, relational, context-embedded practices to solve relational problems" [2]. However, the over reliance on GPS and other abstract mapmaking technologies sometimes makes us to ignore the most obvious human navigational skills.

Indoor navigation without reliable GPS signals is a special problem by itself. Whereas satellite-based systems dominate outdoor navigation, indoors navigation will rely on more powerful wall-mounted antennas, WiFi, and quantum sensors, which will give us our position with an accuracy of centimeters [1]. Underlying these navigation systems is the dismissal of the human wayfinding ability based on visual recognition of spatial features. In the 1950s, Kevin Lynch [3] used simple diagrams to synthesize how residents of Jersey City, Los Angeles, and Boston selected paths and identified borders or landmarks while positioning themselves and finding their way through their cities. Since then, researchers have been capturing affective values attributed to portions of space based on the drawings, photographs, and personal journals of residents. And also, the infinite amount of geotagged digital photographs available online. Using photos posted on social media, researchers have illustrated how different social groups (for example, tourists and residents) experience the same city, and by combining the spatial tags with the cyclical or episodic tags of the photo taken in different cities, researchers find visual signatures of each city [4,5]. Computer-vision techniques have also been utilized to identify spatial features, such as mobile visual location recognition (MVLR), that recognized a landmark based on query image taken by mobile devices, and find more information about it in similar database images [6].

## 1.2 Problem Description

Most of the current state-of-the-art techniques for indoor localization and navigation, including but not limited to signal-based and appearance-based approach, mostly rely on hand-crafted features [7–11]. The performance largely depends on how efficient the hand-crafted feature extractors are developed and to what context it may apply.

In this paper, we propose a method using Deep Convolutional Neural Network (DCNN) to interpret and recognize interior space. DCNN combines deep learning with convolutional neural networks, using millions of free parameters, which are optimized through an extensive training phase. Compared to the hand-crafted features, DCNN combines a powerful ability in feature learning and deep feature extraction, has been tested to learn the best spatial features automatically, which will be robust and relatively universal to various context [12,13]. DCNN has achieved outstanding success in object and scene recognition. Actually, the task to recognize object and scene have a lot in common, which are distinguished from each other based on their inherent physical structure and color pattern. However, how good the ability of DCNN to interpret and recognize a segment of space requires more research. Considering the huge spatial variation it may exist along even in 10 meter-wide indoors corridors, an important research aspect is to understand what spatial features DCNN use in order to gain ability in indoor place recognition.

In this study, we propose a system for robust indoor place recognition using DCNN. In order to test DCNN to recognize location based on spatial features, we focus on a quite bland space: the indoors corridors and atriums that connects several of MIT buildings - the so-called "infinite corridor". A multi-scale indoor place recognition model is introduced to improve the spatial resolution and accuracy. In the application phase, our system, DeepSpace, takes a single photo picture from cellphone as input, and recognize its corresponding pre-zoned location. The preliminary results on the test data set show top-1[1] and top-5[2] error rates of 81.72% and 94.39% respectively. A key feature of our approach for pedestrian indoor navigation is that it does not depend on internet service or special sensors. We focus on the photo analysis from common smartphone cameras, and recognize spatial features of the scene by deep learning, which is more parsimonious and robust.

Different from other methods of wayfinding and spatial location, DCNN interprets space based on visual elements, somehow incorporating the human ability of spatial navigation using visual tips [14]. Indeed, assessing eight state-of-the-art DCNN, [14] used 2D and 3D images, and compared object recognition rates with humans. Based on this important result, which potentially built a connection between DCNN and human cognition, it is also a promising way to understand how human recognize the complex indoor space, by analyzing the spatial features that DCNN learned. Results were promising, showing that DCNN achieved high human-level accuracy when objects are placed against uniform gray backgrounds. However, all those images are outdoors with clearly defined objects outstanding from the background.

The remainder part of this article is organized as follows. Section 1.3 presents related research. In section 2 we introduce the framework of DeepSpace system, which implements the DCNN based indoor navigation, while Section 3 presents the indoor places recognition experiment that we conducted at MIT campus. Section 4 to 5 report on the evaluation, and discuss the results towards deep feature representation and visual similarity of indoor space. Finally, we draw conclusions and present future work in Section 6.

---

[1] The top-1 error rate is the fraction of test images for which the correct label is exactly the label considered most probable by the model.
[2] The top-5 error rate is the fraction of test images for which the correct label is not among the five labels considered most probable by the model.

## 1.3 Related work

In the field of indoor navigation. Legge [15] system for visually-impaired people is an example of indoors wayfinding system that uses handheld sign-reader based on an infrared camera to read digital tags distributed in space, using synthetic speech for output to give simple geometrical descriptions. This is an example of indoors navigation system that use active badges [16], whereas others based on 'Navigation by Scene Familiarity Hypothesis' (NSFH) [17,18], which is one of Insect-inspired navigation algorithm, are also considered a feasible and parsimonious approach for navigation system design.

In the field of indoor localization using computer vision, there are basically two approaches to implement localization: Metric SLAM (Simultaneous localization and mapping) based approach and appearance based approach. SLAM is a popular way to implement localization technique. It focusses on using various types of sensors to build a signal (ultrasonic, WiFi, geomagnetic, bluetooth) map for the target space [19,20]. The resolution and accuracy is high. However, SLAM is computationally expensive due to the complexity 3-D reconstruction and it very much relies on the large hand-crafted feature database and efficient retrieval method. Appearance based localization has achieved good performance in coarsely locating the image among a predefined, limited set of place labels. FAB-MAP [9] introduced a probabilistic approach to recognize places based on appearance information. It built a generative bag-of-words model to represent the discrete scene class.

Recent work in computer vision has proved that using deep convolutional networks has been an effective approach to perform object detection and scene classification. There is a new research trend on using DCNN models to recognize places. Compared with traditional method using images for indoor localization, deep learning has great advantage at feature learning, which have been proved feasible and effective than hand-crafted features. The importance of deep convolutional neural network in support of visual recognition tasks [21,22] has been well recognized with impressive performance.

Kim and Chen [23] trained a DCNN model to learn a control strategy, in order to navigate robot in indoor scenario. Chen, Qu et al.'s DCNN model [23,24] is also used for robot indoor navigation. Differently, the model focus on detecting doors and predicting door's pose. PoseNet [25] combined the strengths of SLAM and DCNN model. They trained a neural network to regress the camera pose from single image, which achieved high resolution and accuracy. Sünderhauf, Shirazi et al. [26] investigated on the utility of DCNN model features for place recognition, and they also evaluate the performance of different level of feature maps to recognize places under severe appearance changes, giving insights into DCNN model's ability to interpret and understand spaces.

Research work in visualizing the features that a model learns in the process of training is also important, in terms of understanding why model performs good and how model be improved. [22] introduced a technique to visualize the activity within the model, helping to identify problems and so obtain better results. Similarly, Class Activation Mapping (CAM) technique [27] allows to highlight the discriminative regions to a specific class, which will be described extensively in section 4.2.1. In this work, we applied CAM to look into model's behavior on how to recognize space and interpret it in an abstract way.

## 2. DeepSpace System: material and methods

In this section we introduce the DeepSpace system framework. The proposed method formulate the indoor localization as a discriminative classification task: Given one image taken from a concerned area, we train a classifier that could predict its corresponding pre-zoned segment. Actually, in order to improve the spatial resolution of localization, based on the low-spatial resolution level segmentation, we further divide each segment into multiple sub-segments, which is considered as high-spatial resolution level, and to train corresponding sub-classifiers. In our case, we divide a sample of MIT campus buildings into 35 segments, to train a building-level DCNN model, and further divide building 7 lobby area into 10 segments, to train a room-level DCNN model. More details is described in experiment section.

Figure1 shows the DeepSpace system framework. In the training phase, we use data extracted from video to train two spatial-scale DCNN models. The first model used data labeled in building-level, which localize the input image to a building segment. The second model is trained with room-level data, aiming at recognizing sub-segment inside one building segment. In service phase, the building-level DCNN model will first predict its building-level segment coarsely, and send to corresponding room-level DCNN model based on its prediction, for high-precision localization.

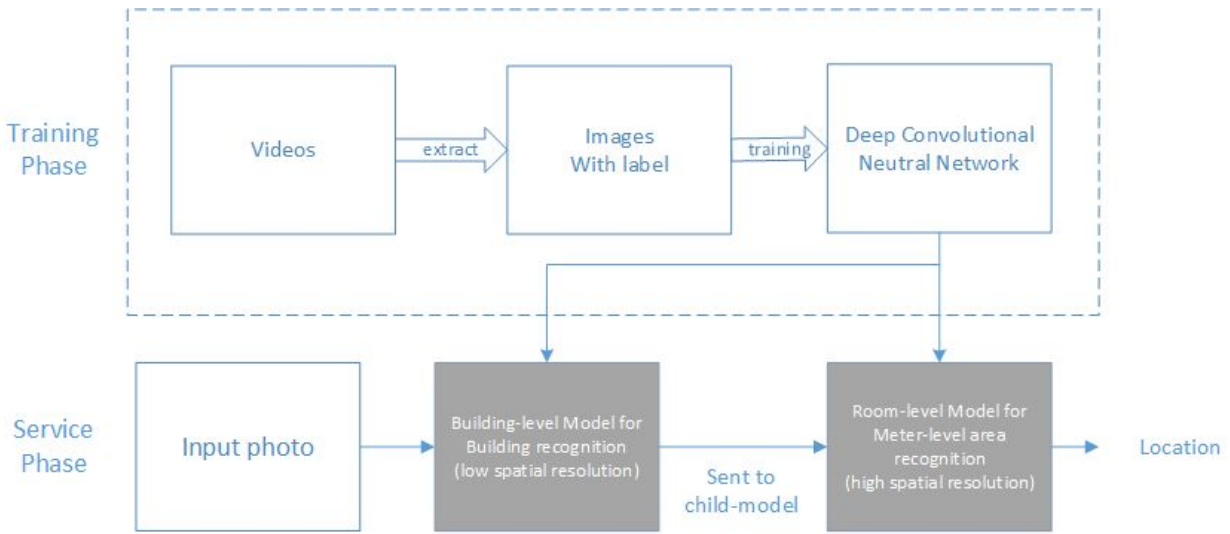

*Figure 1 DeepSpace System Framework. In the training phase, DCNN model is trained with images extracted from video. In service phase, the building-level DCNN model will first predict its building-level segment, and send to room-level DCNN model based on its prediction.*

### 2.1 DeepSpace – the DCNN architecture

We designed our DCNN model mainly with reference to two popular models: AlexNet and NIN [21,28]. Considering the limited amount of images, special attention should be paid to avoid overfitting and to enhance model's generalization capability [29]. As depicted in Figure 2, the network contains 7 weighted layers; the first 3 are Mlpconv layers, in order to learn non-linear features, which will be elaborated below; followed by 2 convolutional layers; and the remaining one is a fully-connected layer. The output of the last fully-connected layer is fed to a 35-way softmax which produces 35 class labels.

The first MLPconv layer filters the 227×227×3 image with 96 kernels, each of size 11×11, using a stride of 4. The resulting feature maps are the pooled (max pooling with 3×3, stride 2) to give 96 corresponding 55 by 55 element feature maps (layer C1). Similarly, the processes are repeated in C2-C5 layers, outputting 512 feature maps with size 11×11. We designed a special pooling layer (P4) – using global average pooling, with same 11×11 window, to deal with the feature map from the last convolutional layer. The last pooling map exports 512 unit features for fully-connected layer (FC). The last layer is a 35-way softmax function, exporting the possibility of the 35 indoor area classes.

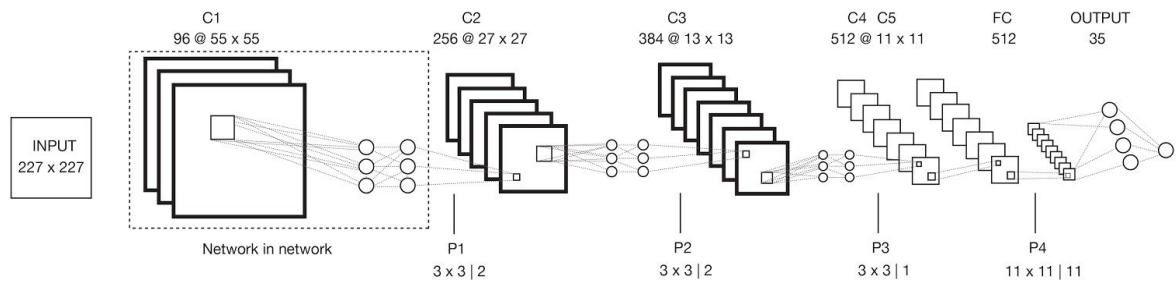

*Figure 2 An illustration of the architecture of DeepSpace - DCNN model. Particularly, the architecture use multilayer perceptron (MLP) convolution layer in the first three convolution layers, and use global average pooling (GAP) followed by a fully-connected softmax layer. The network take a 227×227×3 image as input and the output is 35 categories (vary from specific case studies).*

Below, we describe some of the unusual features of our DCNN model's architecture.

**MLPconv layers**

The classic convolution layer is a Generalized Linear Model (GLM), whose ability of abstraction is believed limited in high-level cognition task [28]. In the same paper, Lin *et al* [28] proposed another structure of convolutional layer – Multi-Layer Perceptron (MLP), which consists of multiple fully connected layers with nonlinear activation functions. By replacing the GLM with this kind of 'micro network' (e.g. MLP), the model is considered to be enhanced in learning non-linear representations. The Figure 3 shows the comparison of classic convolutional layer with Mlpconv layer:

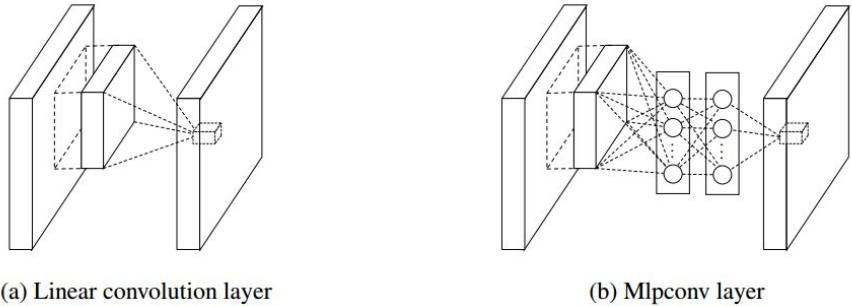

*Figure 3 The classic convolution layer includes a linear filter, whereas Mlpconv consists of a micro neural network, which performs as a nonlinear filter.*

In our DCNN model design, considering the potential of non-linear feature of indoor space, we use Mlpconv as the first three convolution layers.

**Global average pooling (GAP)**

The designed model perform global average pooling at P4 (Figure 3) over the last convolution layer, for two reasons: 1) Recent studies [28] have shown the global average pooling acts as a structural regularizer, which is less prone to overfitting when compared to fully connected layers; 2) GAP is a key structure to implement class activation mapping technology. GAP could retain model localization ability to identify the discriminative image regions. In our case, we used CAM technology to localize the discriminative regions in the image, in order to see what spatial features the model interested in to perform indoor space recognition. Details of CAM technology is described in section 4.

**Other features**

**Dropout** – dropout was first proposed in [21]. It randomly sets half of the activation from feature map to zero during training, which considerably reduces overfitting and improves the model's generalization ability to a large extent. We apply dropout after the first two pooling layers.

**Local Response Normalization (LRN)** – LRN normalizes around the local neighborhood of the unbounded activated neuron, which is considered been beneficial to generalization. The LRN layer performs a sort of lateral inhibition by normalizing over local input regions [21,30], which has been proven to be effective when dealing with Rectified Linear Units (ReLUs).

## 3. Recognizing Indoor Places at MIT Campus

In this section we describe our experiment based on the DeepSpace indoor place recognition system. The experiment was conducted at MIT campus buildings.

## 3.1 Data Collection

MIT Infinite Corridor is the hallway that runs through the main buildings on the campus of the Massachusetts Institute of Technology, in Cambridge, Massachusetts. The Infinite Corridor also serves as the most direct indoor route between the east and west ends of the campus. However, people new to MIT, being students or visitors, usually have difficulty in finding their way when navigating the campus through its hallways.

Figure 4 shows the map of MIT campus buildings, which consist of the building-level research area. Based on the maps of MIT's tunnels and corridors, we divide the ground-level indoors connections among the 26 buildings into 35 segments (with gray color), being 29 segments of corridors and 6 of lobby areas, ranging from 30 $m^2$ to 500 $m^2$.

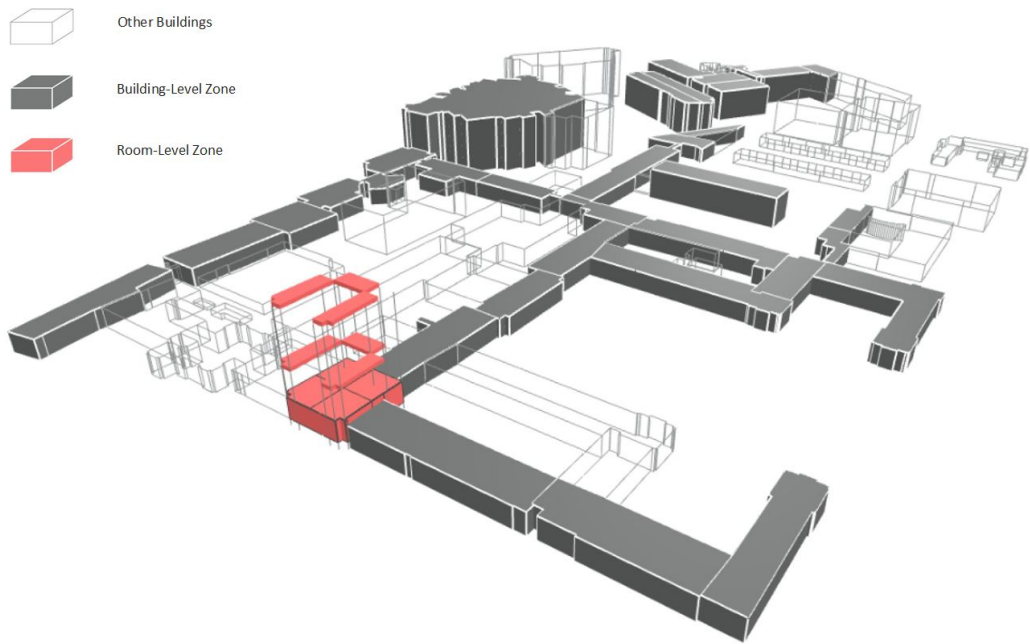

*Figure 4 MIT Campus Buildings Map. 35 indoor segments from 26 Buildings are included in this experiment. The buildings in gray represents buildings involved in the building-level experiment. The building in red shows the area been subdivided into ten room-level areas, in order to increase the spatial resolution for this task. The transparent buildings are irrelevant in experiment.*

For the room-level research area, as is shown in Figure 4 with red color, we took the lobby of building 7 as test site, and divided the area in 10 segments: 4 segments in the 1$^{st}$ floor, 3 segments each in the 2$^{nd}$ and 3$^{rd}$ floors.

We took videos using a commercial action camera GoPro[3] (Figure 5), which generates more stable, clear and high definition video data. Considering the intensity variance of ambient light and the distribution of crowds, we collected several sets of data in different times of day (noon peak hour, afternoon). The device was hand held or fixed on the forehead to keep the camera's shooting angle at the human level. We also tried to cover a wide range of view angles in each position, in order produced a detailed and complete descriptive dataset of indoor spaces.

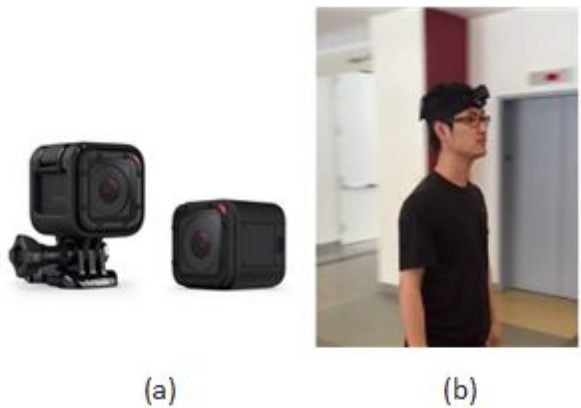

*Figure 5 (a) Action Camera GoPro; (b) Data collection with GoPro fixed on the forehead*

For the building-level data collection, the experimenter made several round-trip walks around each segment while shooting. The number and length of the videos vary with the segment's complexity and area. The video parameter is 720p@30fps. 102 videos are recorded in this section. For the room-level data collection, the experimenter stood in the centric position of each area, shooting the space all around, covering all angles of view. For the test dataset, we collect 1,697 images in total, using a common cell phone (iPhone 6), which cover all the experimental area. In short, the training dataset used GoPro images, and the test dataset used simple cellphone images, which is more close to real usage scenario case.

The images for DCNN training were extracted from videos. The whole procedure includes five steps: 1) Video fisheye distortion correction for all videos, and convert video format to 480p@30fps. This was implemented using GoPro Studio software; 2) Extract images from all frames of the video and rescale to 256×256; 3) Remove low quality images with blur indicator less than 0.45. Here we implement the detecting algorithm refer to [31]; 4) Randomly split the dataset into two: 300 images from each category as validation dataset, and use the rest as training dataset; 5) Generate inventory text file according to the required format of Caffe [32], which is a deep learning framework developed by the Berkeley Vision and Learning Center . Figure 6 shows examples of the image dataset, and the statistics of all datasets is shown in the Figure 7. For segment 'Building X_YZ', number X refers to the building number, character Y indicates the type of segment, in this case, 'C' refers to corridor and 'L' refers to lobby, and Z indicates the serial number of segment in same building. The image amount varies from categories, which is consist with the area and complexity of target segment.

---

[3] GoPro is a copyright holder of GoPro, Inc.

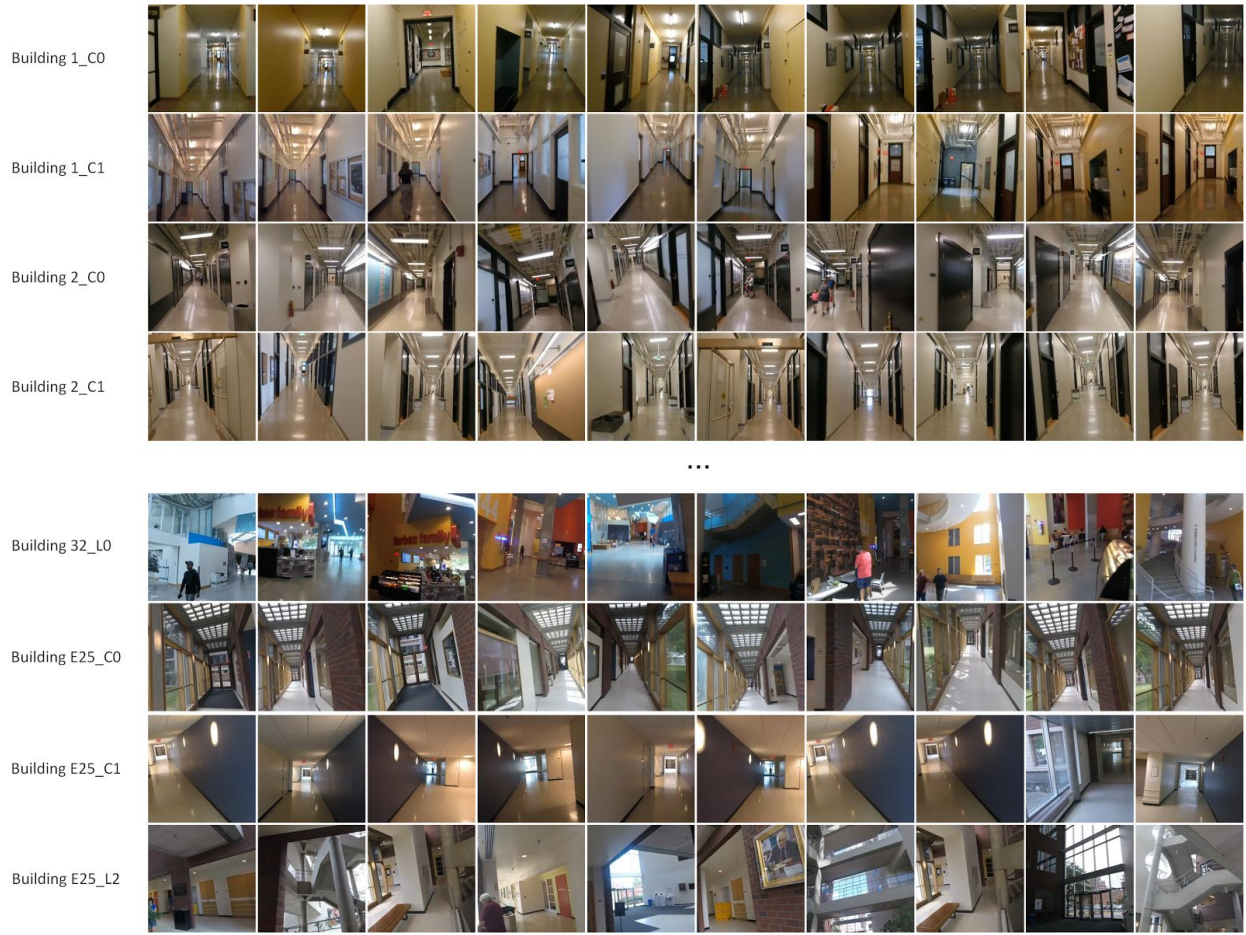

*Figure 6 Example of training dataset. Ten images in each building subzone are randomly selected and shown above. Among them, the type of building 32_L0 and building E52_L2 are lobby, and the remaining ones are corridor.*

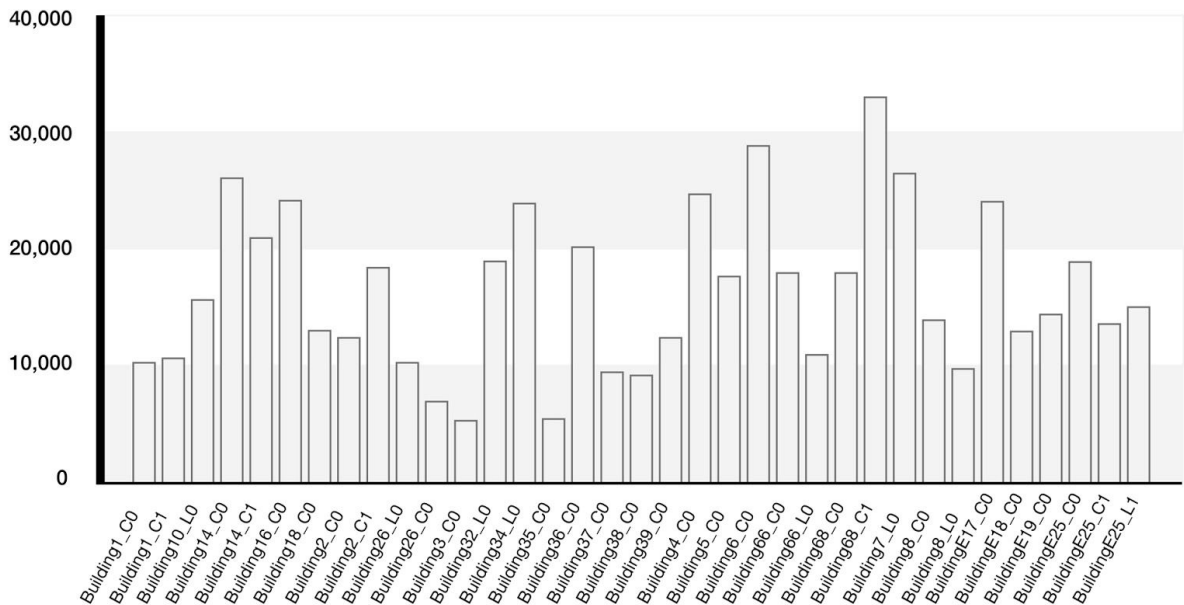

*Figure 7 Statistics of training dataset. A total number of 623,939 images were collected in this experiment.*

## 3.2 Experiment and Preliminary Results

In this section, we separately trained our DCNN model for two tasks: building-level place recognition and building 7 room-level place recognition, on its corresponding dataset.

The DeepSpace DCNN was implemented based on Caffe framework. It was trained using stochastic gradient descent with a base learning rate of 10-4, reduced by 50% every 2000 iterations and with momentum of 0.9. The model is deployed on Amazon Web Service with a g2.2xlarge type instance, which is a parallel computing environment. The server includes virtual GPU computing resource with 1536 CUDA cores and 4GB RAM for video memory. The training process takes 4 hours.

As comparison, on building-level dataset we also evaluated the performance of two other classic DCNNs: AlexNet and NIN, on our dataset. They were both designed for object recognition. AlexNet won the first place in ILSVRC2012 (ImageNet Large Scale Visual Recognition Challenge 2012) competition, showing remarkable performances on extensively large and difficult object databases. NIN, with a small size of parameters, performs slightly better than AlexNet on ImageNet dataset. For room-level place recognition, we only tested DeepSpace to see if the model is also effective in small spatial scale dataset.

The performance of the three DCNN models on our MIT Campus building-level dataset is compared in Table 1. On the training and validation dataset, the three networks achieved almost the same accuracy. In fact, AlexNet performed slightly better, at the cost of ten times of model size than other two. On the test dataset, DeepSpace achieves significantly higher accuracy than other two models, in particular, with 81.72% in top-1 accuracy. At room-level location, we only trained DeepSpace on the room-level dataset, which achieved 96.90% top-1 accuracy on validation dataset.

*Table 1: Performance comparison between AlexNet, NIN and our designed model on building-level dataset.*

| Model | Accuracy Top-1 Validation | Accuracy Top-1 Test | Accuracy Top-5 Test |
| --- | --- | --- | --- |
| *AlexNet* | 99.80% | 48.35% | 80.90% |
| *NIN* | 98.10% | 59.19% | 89.33% |
| *DeepSpace* | **97.90%** | **81.72%** | **94.39%** |

Examples of recognition result are shown in Figure 8. Based on given images, the model predicts their location with confidence. We use class activating mapping (CAM) technology to highlight the discriminative image regions. The informative regions will help us to know how DCNN interprets and recognizes the complex interior space (details described in section 5).

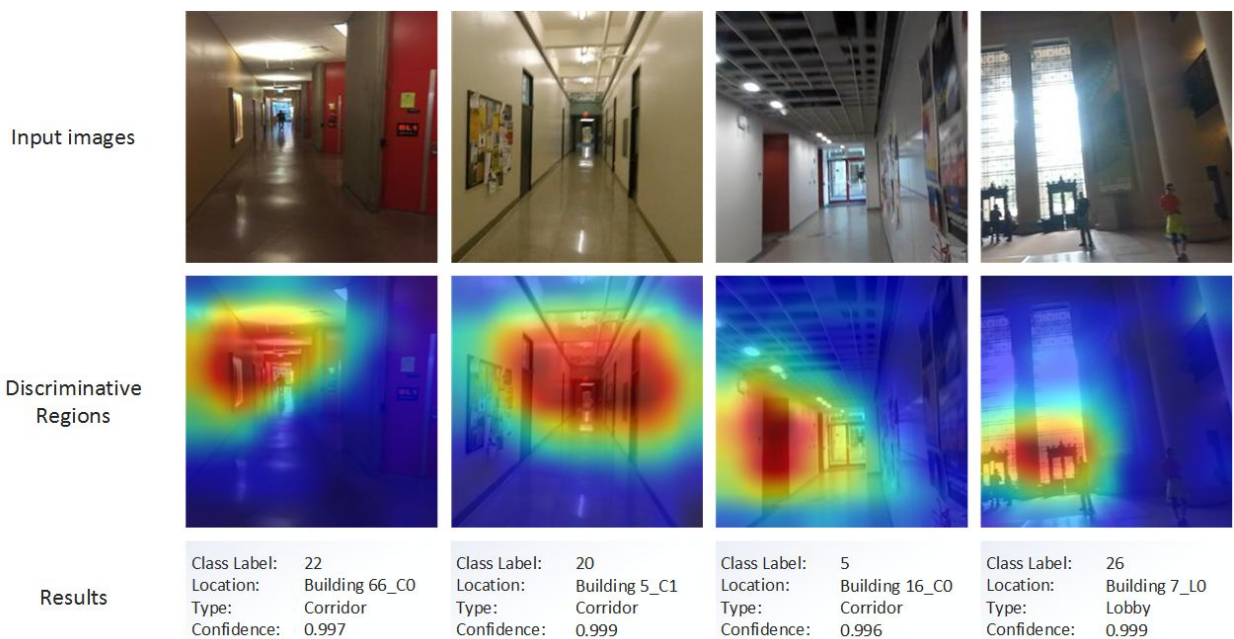

*Figure 8 Examples of indoor place recognition inside MIT Buildings. We use Class Activating Mapping (CAM) to highlight the discriminative image regions.*

## 4. Deep Feature Representations of interior Space

In this section we look into the interior space image's deep representations that learned by the DCNN model. Considering the huge spatial variation it may exist even in 10 meter-wide indoors corridors, an important research aspect is to understand what spatial features DCNN relies on, in order to acquire the ability in indoor place recognition task. In this point, we mainly focus on analyzing what DCNN model learned and how DCNN model recognize and interpret interior space. To realize this, the two approaches were:

## 4.1 Filter map analysis

We first visualized the filter map of network's first convolutional layer. The filter map is used for filtering the raw image data at first step. Basically, it shows the model's 'philosophy' about what kind of elements that constitute the image from a specific category.

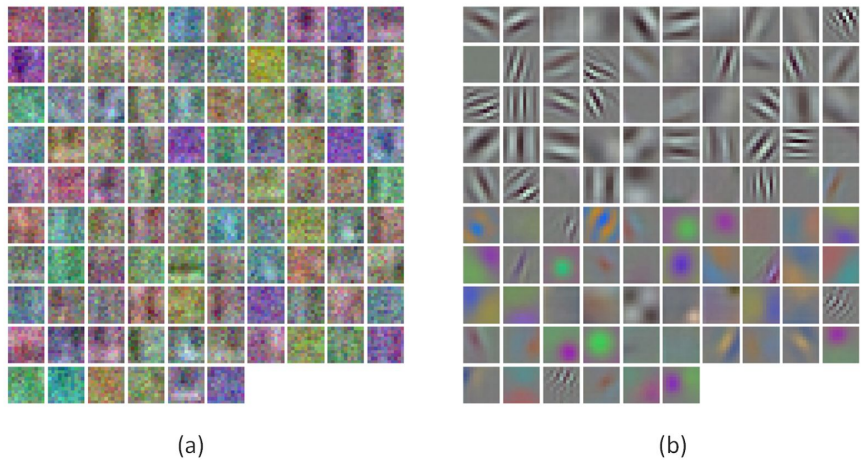

(a)                                                                 (b)

*Figure 9 96 convolutional kernels of size 11×11×3 learned by the 1st convolutional layer on the 227×227×3 input images. (a) Kernels learned in space recognition task (this study); (b) Kernels learned in ImageNet Classification task*

Figure 9 shows the filter map of 1st convolutional layer from two different classification tasks. Figure 9 (b) is well recognized as a classic example to show DCNN model's ability of feature extraction. The filter map in figure 9 (b) is learnt from a very general and common task, the ImageNet Classification task, which is to classify 1000 categories of objects and scenes. The network learns all kinds of frequency, orientation and color sensitive kernels. Compared with our results, as shown in figure 9 (a), the interior space classification task learned quite different filters, which mostly focuses on color patterns. The reason for this, as far as we can infer, is that the ImageNet task is object-detection oriented. The object in same category exists invariant visual representation in shape, which could be linearly represented by contour and activate frequency—or orientation-sensitive neurons. While for the space classification task, model is not based on single objects, scenes, or even fixed spatial shapes, which would not be sensitive to frequency and orientation based filters. That makes sense, for it is in line with human's perception and localization ability in interior space. We recognize place inside a building based on our holistic perception, combining the color and patterns of decoration, the pattern of spatial features, what is quite different from a single object recognition.

## 4.2 Understanding model's behavior

We are also interested in whether there are image-level representations for the DCNN model. We tried to localize discriminative image regions for this indoor image classification task, to find out what visual spatial properties that the model is interested in. As a first step, we introduce the Class Activation Mapping (CAM) method.

**Class Activation Mapping**

Recent works in DCNN model feature visualization have shown that the model retains a remarkable ability to localize objects in the convolutional layers until the final fully-connected layer, which increases the potential to identify the discriminative image region in a single forward pass process. [27] suggests a feasible way to implement this idea. Combining our defined model architecture, we describe the detailed procedure in terms of CAM technique:

In a forward pass, we define $F_k(x,y)$ as the last convolutional layer's feature map $k$, and define $\omega$ as the weight vector between global average pooling layer and softmax layer. According to the final classification result, we will have the best matching class $c$ and its corresponding weight $\omega_c$. Here we take $\omega_c$ as the best linear combination weights for feature map $F_k(x,y)$, in terms of obtaining the importance of the activation at feature map space. Hence, the class activation map can be given by:

$$M_c = \sum_k \omega_c F_k(x,y)$$

The size of $M_c$ would be the same as the last convolutional layer, in our case, 11 by 11. We just need to upsample it to 227 by 227, and composite the original input image. We can identify the images' regions most relevant to the class, to further understand the visual spatial feature that the model is most interested in.

**Visual Element Identification**

The following part applied CAM technique to identify the informative region. From our dataset we found that the scene varies in a large range, even in a same building zone. Unlike image data for object detection, which depicted objects from different angles of view, indoor space classification task doesn't rely on single-object level recognition [33]. The common ground for a segment of space consisted of global spatial features and high-level visual features. They are more like style and pattern of spatial properties.

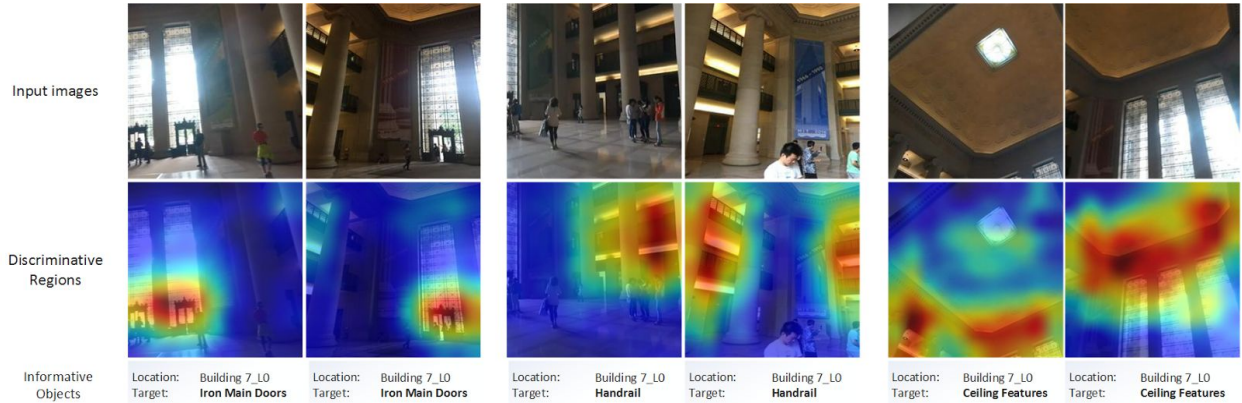

*Figure 9 6 images from same indoor location - the lobby of building 7. The discriminative region highlights the most informative objects or patterns considered by the model. They vary a lot even in same categories.*

Figure 9 shows 6 examples of input image with discriminative region. The 6 images are all from a same segment– the lobby of building 7. Images in this region are recognized with high accuracy by the model, but the discriminative spatial features vary a lot. The first two images were identified by the iron main doors; the middle two images were identified by the handrail; and the last two images were identified by ceiling features. This shows that the model recognizes indoor space based on multiple, not on a single object.

## 5. Visual Similarity of Indoor Space

How visually similar or different are these buildings? Based on human perception, we may recognize the decoration of some corridors in west campus has a same style. But how to measure the similarity in a quantitative way is worth studying.

### 5.1 What pair of indoor places look similar

In this part we refer to Bolei's research [34] in measuring the similarity of cities based on misclassification rate, which assume that if images in two categories are visually similar, the misclassification rate in the recognition task will be high. In our case we use normalized misclassification matrix to identity the visually similar pair among MIT Campus buildings.

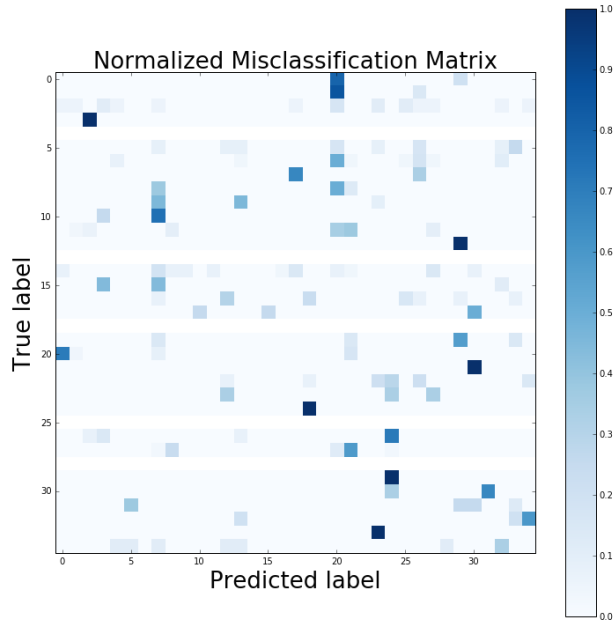

*Figure 10 Normalized Misclassification Matrix. Each cell (i, j) in the matrix indicate the extent to which category i is misclassified as category j.*

Figure 10 shows the normalized misclassification matrix, which is the confusion matrix with true positive value removed. The misclassification rate on the pair - indoor place (A, B), is computed by the sum of the rate of misclassifying images of place A as place B and the rate of misclassifying images of place B as place A. Based on this calculation, we rank all the possible pairwise indoor places, and drew similarity graph of MIT Campus buildings, as is shown in Figure 11. The thickness of the edges, which is based on the misclassification rate in misclassification matrix, indicates the visual similarity between pairs of buildings. From this figure we can see that the buildings being next to each other tend to be more similar, and also buildings distributed around West Campus tend to be more similar to each other.

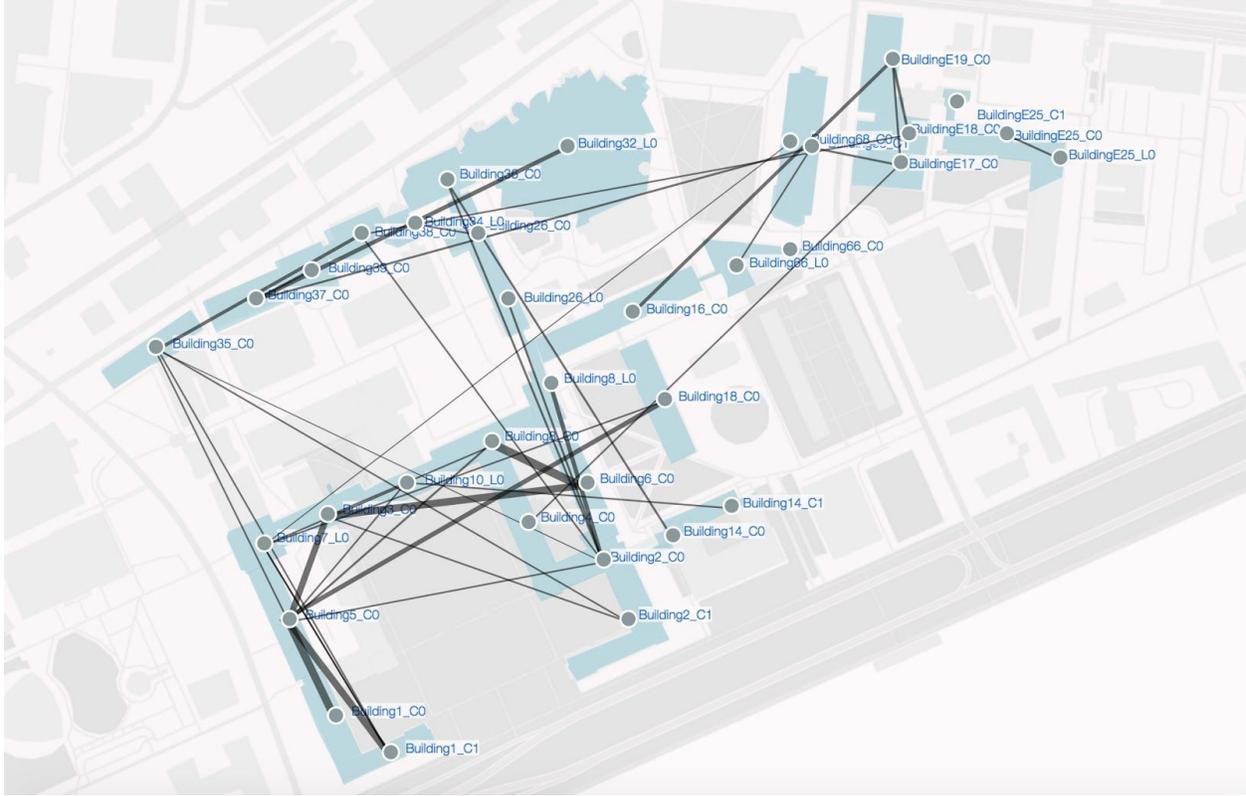

*Figure 11. Similarity Graph of MIT Campus Buildings. The thickness of the edges, which is based on the FN score in misclassification matrix, indicates the visual similarity between pairs of buildings.*

## 5.2 What are the most distinctive indoor places at MIT

Furthermore, based on the misclassification matrix, we designed an evaluation method to assess how distinctive an indoor place is. In the misclassification matrix, we focus on two factors: the rate of false positive (FP), which refer to one category that all other classes were incorrectly labeled as; and also the rate of false negative (FN), corresponding to one category that was incorrectly marked as pertaining to any other classes. To some extent, false positive and false negative reflects the similarity and distinctiveness of one category. In this case, we simply take the sum of these two factors, to show how distinctive that one segment space is.

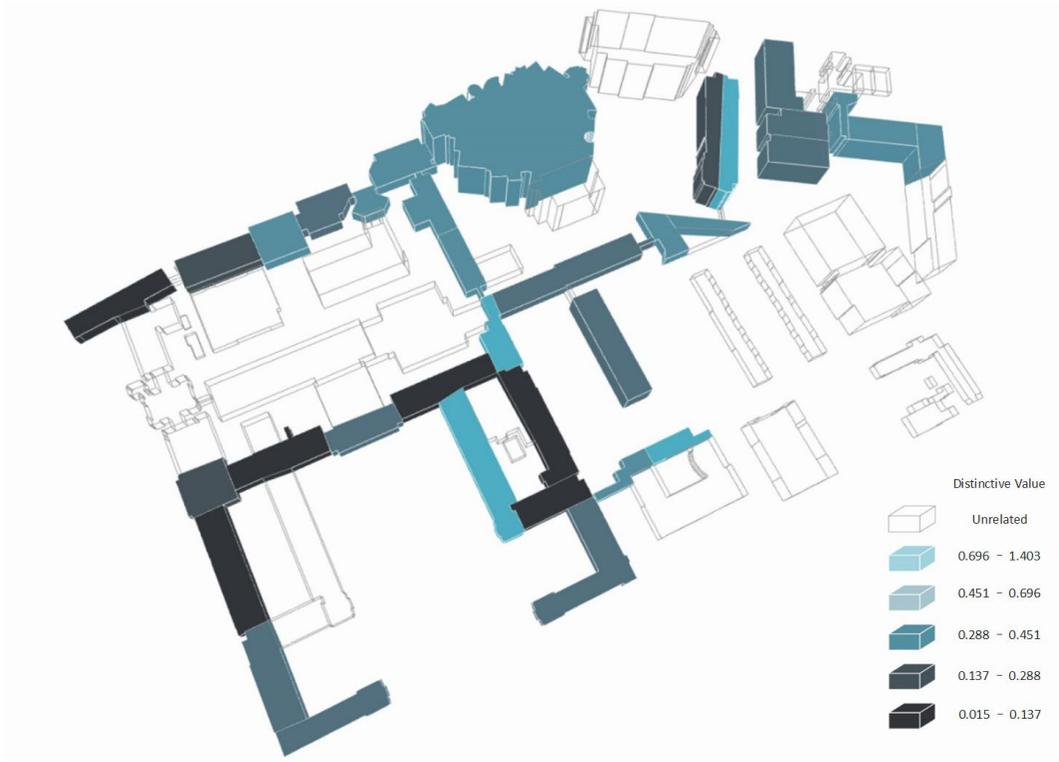

*Figure 12 Distinctive map of MIT Campus Buildings. The buildings with brighter color are considered to be more distinctive.*

Figure 12 shows the distinctive map of buildings at MIT campus. We can see the west campus, which is shown on the left of the map, are significantly with lower distinctive score. One outlier happens in building 4_C0, which is considered as a distinctive area, surrounded by all other building which is visually 'not distinctive'. One explanation for this outlier is that building 4 is actually undergoing renovation, with half of the corridors occupied by equipment and scaffold, which makes the area visually distinctive. Beyond that, building 8_L1, building 14_C1, building 68_C1, building 34_L0 are considered as most distinctive area. Example images are shown in figure 13.

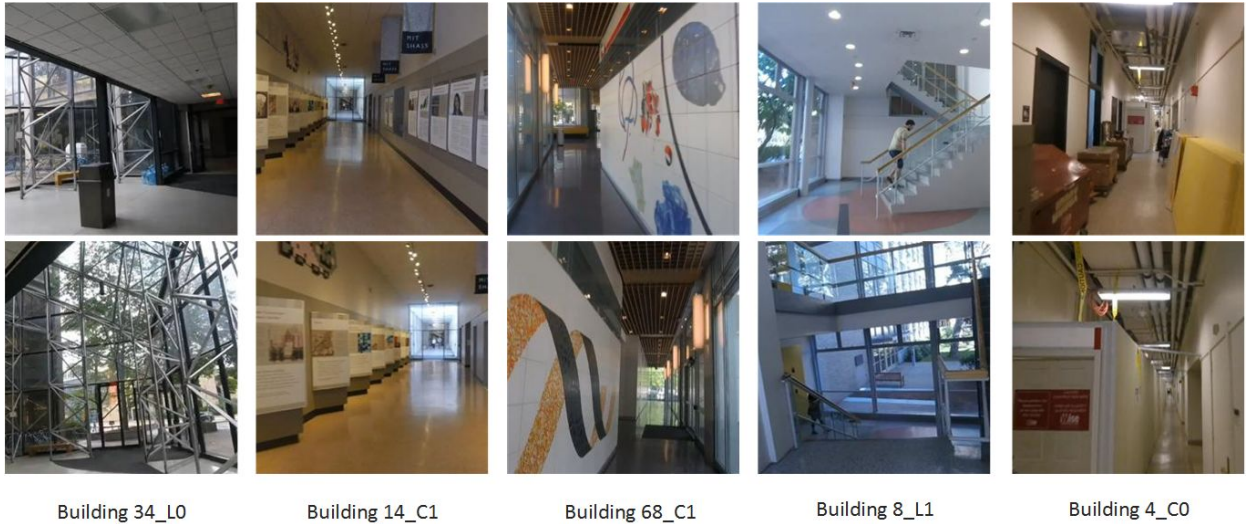

*Figure 13 Most distinctive area at MIT Campus, of which, building 4_C0 is undergoing renovation.*

## 6. Discussion and future work

Both outdoor and indoor navigation technologies have become increasingly accurate, either relying on GPS coverage, or Wi-Fi and other communication technologies. However, in this process the key human future of navigating space based on spatial recognition based on visual cues, is not taken into account. More recently, computer vision, machine learning and other more advanced techniques have been tested to taken into account visual features as a way of identifying particular spaces, and increment spatial navigation.

In this paper we introduced a robust and parsimonious indoor place recognition method using a DCNN model, which achieved good results in a case study realized at MIT campus (at a high accuracy). We present a multis-cale hierarchical model to improve the resolution and accuracy of localization. The model is parsimonious without having the aid of vision- or signal-based handcrafted features, and the method could be adapted to various scenarios. The results show that DCNN model has the ability to interpret the space in a comprehensive way and recognize interior space with high accuracy, which could be applied to indoor localization and navigation.

From a modeling perspective, the DCNN model learns the inherent features that distinguish the indoor space with one another, providing insight to the study of interior space, to design a more targeted DCNN architecture which could perform better in large-scale indoor localization. Future works will continue to focus on analyzing the deep features that DCNN learns, understanding model's behavior in decision process, and somehow comparing with human's indoor perception and wayfind ability. By understanding the relationship between spatial features and confusion degree, the long term significance of future works aims to bring insights in how to conduct architectural design using friendlier approaches, from the perspective of human cognitive architecture, enhancing people's recognition of spatial features and navigation in indoor environments.

# Supporting Information

**S1 Dataset. The Dataset includes 623,939 images used for training, validating and testing indoor place recognition models.** The images were extracted from 102 raw videos taken inside 26 MIT campus buildings, including 35 segments (categories). The size of image is 255 by 255, and the format is JPEG.

# Acknowledgements

This work was supported by the National Key Basic Research Program of China (2015CB954103), the National Natural Science Foundation of China (grant no. 41371388) . This work was also supported by the Hong Kong Research Grants Council (RGC) General Research Fund (GRF) project 14606715.

Authors wish to thank the MIT SENSEable City Lab Consortium.

# Author Contributions

Conceived and designed the experiments: FZ RM. Performed the experiments: FZ. Analyzed the data: FZ FD. Contributed reagents/materials/analysis tools: FZ RM. Wrote the paper: FZ FD. Original Idea: FZ RC DM. Supervision: HL RC.

# Reference

1. McKinlay R, Roger M. Technology: Use or lose our navigation skills. Nature. 2016;531: 573–575.

2. Kitchin R, Dodge M. Rethinking maps. Prog Hum Geogr. 2007;31: 331–344.

3. Lynch K. The Image of the City. MIT Press; 1960.

4. Luo J, Jiebo L, Dhiraj J, Jie Y, Andrew G. Geotagging in multimedia and computer vision—a survey. Multimed Tools Appl. 2010;51: 187–211.

5. Paldino S, Silvia P, Iva B, Stanislav S, Carlo R, González MC. Urban magnetism through the lens of geo-tagged photography. EPJ Data Science. 2015;4. doi:10.1140/epjds/s13688-015-0043-3

6. Guan T, Fan Y, Duan L, Yu J. On-device mobile visual location recognition by using panoramic images and compressed sensing based visual descriptors. PLoS One. 2014;9: e98806.

7. Wolf J, Burgard W, Burkhardt H. Robust vision-based localization by combining an image-retrieval system with Monte Carlo localization. IEEE Trans Rob. ieeexplore.ieee.org; 2005;21: 208–216.

8. Filliat D, Meyer J-A. Map-based navigation in mobile robots:: I. A review of localization strategies. Cogn Syst Res. 2003;4: 243–282.

9. Cummins M, Newman P. FAB-MAP: Probabilistic localization and mapping in the space of appearance. Int J Rob Res. ijr.sagepub.com; 2008; Available:

http://ijr.sagepub.com/content/27/6/647.short

10. Mirowski P, Milioris D, Whiting P, Kam Ho T. Probabilistic Radio-Frequency Fingerprinting and Localization on the Run. Bell Labs Tech J. New York, NY, USA: John Wiley & Sons, Inc.; 2014;18: 111–133.

11. Milioris D, Tzagkarakis G, Papakonstantinou A, Papadopouli M, Tsakalides P. Low-dimensional signal-strength fingerprint-based positioning in wireless LANs. Ad Hoc Networks. 2014/1;12: 100–114.

12. Donahue J, Jia Y, Vinyals O, Hoffman J, Zhang N, Tzeng E, et al. DeCAF: A Deep Convolutional Activation Feature for Generic Visual Recognition. Proceedings of The 31st International Conference on Machine Learning. jmlr.org; 2014. pp. 647–655.

13. Oquab M, Bottou L, Laptev I, Sivic J. Learning and transferring mid-level image representations using convolutional neural networks. Proc IEEE. cv-foundation.org; 2014; Available: http://www.cv-foundation.org/openaccess/content_cvpr_2014/html/Oquab_Learning_and_Transferring_2014_CVPR_paper.html

14. Kheradpisheh SR, Ghodrati M, Ganjtabesh M, Masquelier T. Deep Networks Can Resemble Human Feed-forward Vision in Invariant Object Recognition. Sci Rep. 2016;6: 32672.

15. Legge GE, Beckmann PJ, Tjan BS, Havey G, Kramer K, Rolkosky D, et al. Indoor navigation by people with visual impairment using a digital sign system. PLoS One. 2013;8: e76783.

16. Hightower J, Borriello G. Location systems for ubiquitous computing. Computer . 2001;34: 57–66.

17. Gaffin DD, Brayfield BP. Autonomous Visual Navigation of an Indoor Environment Using a Parsimonious, Insect Inspired Familiarity Algorithm. PLoS One. Public Library of Science; 2016;11: e0153706.

18. Gaffin DD, Dewar A, Graham P, Philippides A. Insect-Inspired Navigation Algorithm for an Aerial Agent Using Satellite Imagery. PLoS One. Public Library of Science; 2015;10: e0122077.

19. Newcombe RA, Lovegrove SJ, Davison AJ. DTAM: Dense tracking and mapping in real-time. 2011 International Conference on Computer Vision. 2011. doi:10.1109/iccv.2011.6126513

20. Engel J, Jakob E, Thomas S, Daniel C. LSD-SLAM: Large-Scale Direct Monocular SLAM. Lecture Notes in Computer Science. 2014. pp. 834–849.

21. Krizhevsky A, Sutskever I, Hinton GE. ImageNet Classification with Deep Convolutional Neural Networks. In: Pereira F, Burges CJC, Bottou L, Weinberger KQ, editors. Advances in Neural Information Processing Systems 25. Curran Associates, Inc.; 2012. pp. 1097–1105.

22. Zeiler MD, Rob F. Visualizing and Understanding Convolutional Networks. Lecture Notes in Computer Science. 2014. pp. 818–833.

23. Kim DK, Chen T. Deep Neural Network for Real-Time Autonomous Indoor Navigation [Internet]. arXiv [cs.CV]. 2015. Available: http://arxiv.org/abs/1511.04668

24. Chen W, Wei C, Ting Q, Yimin Z, Kaijian W, Gang W, et al. Door recognition and deep learning

algorithm for visual based robot navigation. 2014 IEEE International Conference on Robotics and Biomimetics (ROBIO 2014). 2014. doi:10.1109/robio.2014.7090595

25. Kendall A, Alex K, Matthew G, Roberto C. PoseNet: A Convolutional Network for Real-Time 6-DOF Camera Relocalization. 2015 IEEE International Conference on Computer Vision (ICCV). 2015. doi:10.1109/iccv.2015.336

26. Sunderhauf N, Niko S, Sareh S, Feras D, Ben U, Michael M. On the performance of ConvNet features for place recognition. 2015 IEEE/RSJ International Conference on Intelligent Robots and Systems (IROS). 2015. doi:10.1109/iros.2015.7353986

27. Zhou B, Khosla A, Lapedriza A, Oliva A, Torralba A. Learning Deep Features for Discriminative Localization [Internet]. arXiv [cs.CV]. 2015. Available: http://arxiv.org/abs/1512.04150

28. Lin M, Chen Q, Yan S. Network In Network [Internet]. arXiv [cs.NE]. 2013. Available: http://arxiv.org/abs/1312.4400

29. Hinton GE, Srivastava N, Krizhevsky A, Sutskever I, Salakhutdinov RR. Improving neural networks by preventing co-adaptation of feature detectors [Internet]. arXiv [cs.NE]. 2012. Available: http://arxiv.org/abs/1207.0580

30. Lyu S, Simoncelli EP. Nonlinear Image Representation Using Divisive Normalization. Proc IEEE Comput Soc Conf Comput Vis Pattern Recognit. 2008;2008: 1–8.

31. Tong H, Li M, Zhang H, Zhang C. Blur detection for digital images using wavelet transform. Multimedia and Expo, 2004 ICME '04 2004 IEEE International Conference on. 2004. pp. 17–20 Vol.1.

32. Jia Y, Shelhamer E, Donahue J, Karayev S, Long J, Girshick R, et al. Caffe: Convolutional Architecture for Fast Feature Embedding. Proceedings of the 22Nd ACM International Conference on Multimedia. New York, NY, USA: ACM; 2014. pp. 675–678.

33. Quattoni A, Torralba A. Recognizing indoor scenes. Computer Vision and Pattern Recognition, 2009 CVPR 2009 IEEE Conference on. ieeexplore.ieee.org; 2009. pp. 413–420.

34. Zhou B, Bolei Z, Liu L, Aude O, Antonio T. Recognizing City Identity via Attribute Analysis of Geo-tagged Images. Lecture Notes in Computer Science. 2014. pp. 519–534.